\documentclass[letterpaper, 10 pt, conference]{ieeeconf}
\IEEEoverridecommandlockouts
\IEEEoverridecommandlockouts                              

\usepackage{svg}
\usepackage{xcolor}
\usepackage{comment}

\usepackage{graphicx} 
\usepackage{subcaption}
\usepackage{stfloats}  
\usepackage{amsmath} 
\usepackage{bm} 
\usepackage{amssymb}  
\usepackage{pifont}   
\usepackage{url}
\usepackage{hyperref}
\usepackage[usenames,dvipsnames,table]{xcolor}

\usepackage{booktabs}
\usepackage{xspace}
\usepackage{multirow}
\usepackage{amsfonts}
\newcommand{\cmark}{\ding{51}} 
\newcommand{\xmark}{\ding{55}} 

\usepackage{makecell}
\usepackage[table]{xcolor}

\makeatletter
    \let\NAT@parse\undefined
\makeatother
\usepackage[square,numbers,sort&compress]{natbib}

\usepackage{hyperref}
\hypersetup{
    colorlinks=true,
    linkcolor=MidnightBlue,
    filecolor=magenta,      
    urlcolor=MidnightBlue,
    citecolor=MidnightBlue,
}
\usepackage[capitalise, nameinlink]{cleveref}

\title{\LARGE \bf
VLA-ACL: Action-Consistent Visual Token Pruning for Efficient Vision-Language-Action Models
}

\author{Owen Du$^{1,2,*}$ Yang Yue$^{2,*}$, Jie Zhang$^{2}$, Jiaqi Pi$^{2}$, Chi Bene Chen$^{2}$, Gao Huang$^{2,\dagger}$
\thanks{$^{1}$ETH Zurich\,\, $^{2}$Tsinghua University}
\thanks{$^*$Equal contribution}
\thanks{$^{\dagger}$Corresponding author}
}

\begin{document}

\maketitle
\thispagestyle{empty}
\pagestyle{empty}

\begin{abstract}
Vision-Language-Action (VLA) models achieve strong robotic manipulation performance but incur high computational costs from processing long token sequences at every control step, limiting real-time deployment.
Visual token pruning offers a direct solution, as visual patches dominate the input sequence and contain considerable redundancy. Existing approaches, however, either rely on indirect training-free heuristics, such as attention scores and motion thresholds, or require costly fine-tuning of the base VLA model.
We introduce VLA-ACL (\textit{Action Consistency Learning}), which learns a lightweight visual token pruning policy through action-level supervision while keeping the base VLA model entirely frozen.
The training objective encourages actions produced from pruned visual contexts to remain consistent with the full-context teacher, with ground-truth actions as auxiliary supervision. This directly ties token selection to its effect on the downstream control output.
Experiments on LIBERO and real-world manipulation tasks show that VLA-ACL prunes up to 87.5\% of visual tokens while retaining competitive performance, reduces computation by up to 75\%, and achieves a 1.5$\times$ inference speedup. These results establish a stronger performance–efficiency trade-off than existing frozen-VLA pruning methods and demonstrate the value of action-level supervision for visual token selection. Code \href{https://github.com/du-owen/VLA-ACL}{available}.
\end{abstract}

\section{Introduction}

Vision-Language-Action (VLA) models combine pretrained vision-language representations with action prediction to achieve strong performance in robotic manipulation~\cite{pmlr-v270-kim25c,kim2025fine,black2024pi_0,Intelligence202505AV}. However, their computational cost limits real-time deployment. At every control step, these models process long multimodal sequences dominated by visual tokens from camera observations. Processing these tokens through a large language backbone incurs substantial computation and latency, constraining the frequency at which the policy can respond to changes in the environment.

One promising direction for closing this gap is to exploit visual redundancy through token pruning or caching. Pruning reduces the number of visual tokens processed, while caching reuses visual representations across control steps.
Prior work~\cite{wang2026specprunevla} shows that VLA policies can tolerate even random token removal at moderate pruning rates, suggesting substantial redundancy in the visual input. The challenge is to exploit this redundancy more aggressively while preserving the visual information needed for effective control.

Existing approaches largely fall into two categories. Training-free methods~\cite{NEURIPS2025_f062da19,ICLR2026_1223ca2b,ma2026safe,liu2025bridging,NEURIPS2025_3a2ef31a,li2025sp,wang2026specprunevla,tan2025think} freeze the base VLA and select tokens using heuristics such as attention scores, temporal changes, and motion thresholds. However, these signals only indirectly estimate token importance and do not capture the actual impact of token removal on model performance. Moreover, such methods often require carefully tuned layer- or timestep-specific hyperparameters or access to internal states and attention maps, limiting compatibility with existing acceleration techniques. Training-based approaches~\cite{jiang2025better,feng2026see} instead learn token selection jointly with the base VLA, incurring substantial backbone fine-tuning costs~\cite{kim2025fine}. This also couples the resulting VLA checkpoint to the specific pruning mechanism, limiting its deployment flexibility. These limitations motivate a central question: \textit{Can we learn effective visual token selection from its impact on downstream actions while keeping the base VLA frozen?}

\begin{figure}[tbp]
\centering
\includegraphics[width=\linewidth]{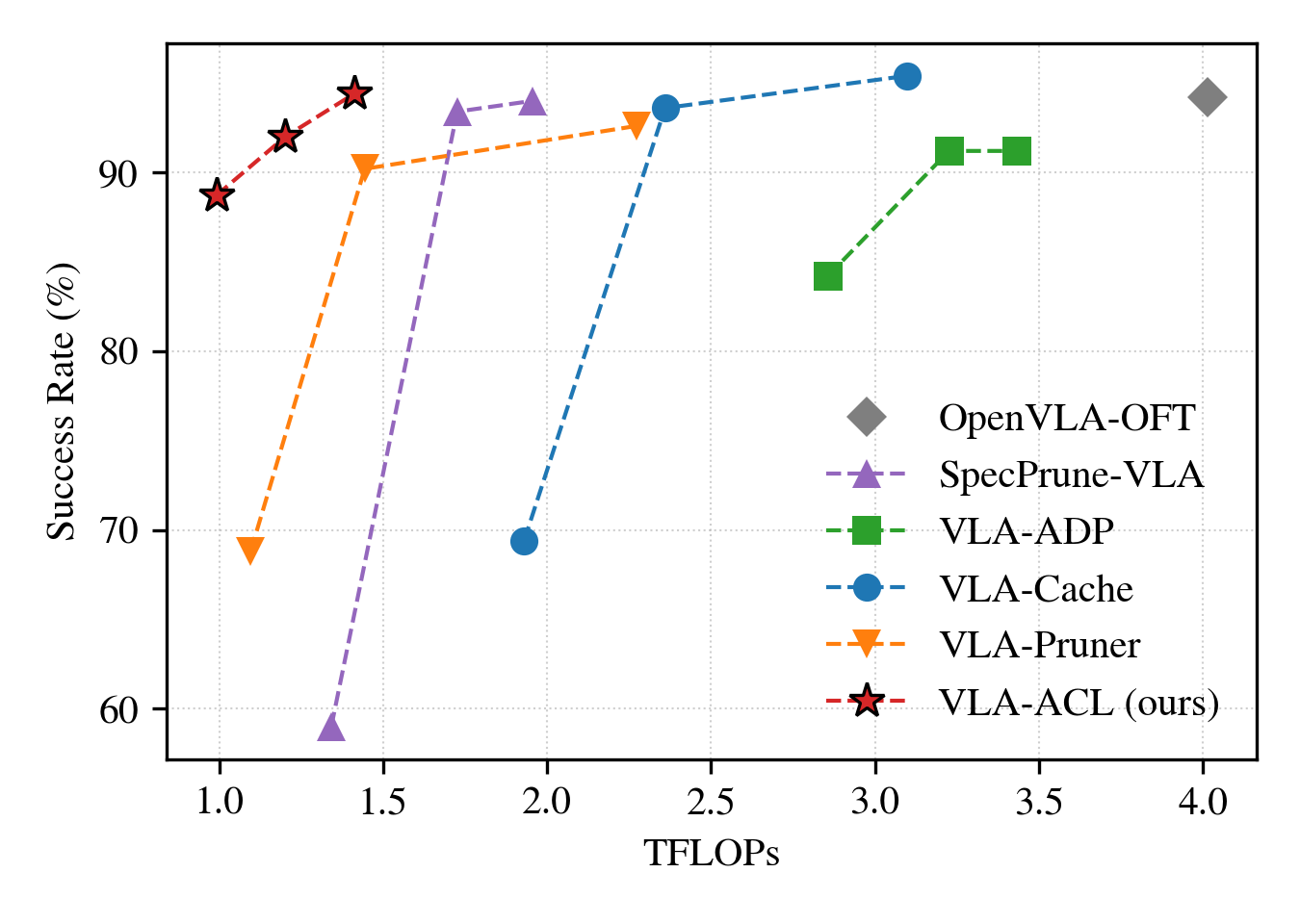}
\caption{LIBERO-Long success rates vs. computational budget. \textbf{VLA-ACL} achieves a better performance-efficiency trade-off than existing frozen-VLA pruning and caching methods.}
\label{fig:libero-10-pareto}
\end{figure}

To this end, we introduce VLA-\textbf{ACL} (\textbf{A}ction \textbf{C}onsistency \textbf{L}earning), a visual pruning framework for VLAs that trains a lightweight token selection policy based on its effect on the predicted action: actions produced from \underline{pruned visual contexts} should remain \textbf{consistent} with those produced by the same VLA using the \underline{full visual context} (Fig.~\ref{fig:full-pipeline}). Combined with auxiliary ground-truth action supervision, this teacher-consistency objective directly links visual token selection to downstream control behavior. VLA-ACL trains a patch scoring network to assign an importance score to each visual token and retains the 
$K$ highest-scoring tokens per camera view before passing them to the language backbone at inference.
During training, to enable gradient flow to the patch scorer, we use a soft Top-$K$ relaxation~\cite{lapsum2025} to gate visual embeddings and gradually sharpen the selection weights, reducing the mismatch with hard pruning at inference. Our framework neither accesses nor modifies the internal states of the base model, making it broadly compatible with existing VLA architectures.

Empirically, VLA-ACL requires just a few hours of training on a single GPU, using only a lightweight Transformer \cite{attention} as the patch scorer, and avoids costly multi-GPU fine-tuning of the full policy. We evaluate VLA-ACL on LIBERO and real-world manipulation tasks. Our framework supports aggressive pruning rates up to 87.5\% of visual tokens and reduces FLOPs by up to 75\%, while maintaining success rates comparable to full-context inference.
These results demonstrate a strong performance–efficiency trade-off against frozen-VLA pruning baselines, as shown in Fig.~\ref{fig:libero-10-pareto} and Table~\ref{tab:main_libero_results}. VLA-ACL thus establishes action consistency as an effective learning signal for visual token selection, enabling aggressive pruning while preserving the information needed for control.

\section{Related Work}

\subsection{Vision-Language-Action Models}

Vision-Language-Action (VLA) models build on pre-trained vision-language representations~\cite{zhai2023sigmoid,oquab2024dinov} to predict robot actions from visual observations and language instructions~\cite{Beyer2024PaliGemmaAV,Karamcheti2024PrismaticVI}. Representative models include OpenVLA~\cite{pmlr-v270-kim25c} and the $\pi$-series~\cite{black2024pi_0,Intelligence202505AV}. OpenVLA processes visual patch tokens and language instructions through a large language backbone~\cite{Touvron2023Llama2O}, while OpenVLA-OFT~\cite{kim2025fine} improves action-generation efficiency through parallel decoding of action chunks. Despite these advances, processing long visual token sequences remains computationally expensive, hindering their practicality for real-world tasks.

\subsection{Efficient VLA Inference}\label{sec:related-efficiency}

VLA inference acceleration has been explored through compact architectures \cite{shukor2025smolvla,Wen2024TinyVLATF}, dynamic depth control \cite{deervla}, and visual token pruning. Among pruning approaches, \textit{training-free} methods operate on a frozen VLA stack and offer plug-and-play inference acceleration. Building on token pruning for vision-language models \cite{chen2024image,zhang2024sparsevlm,alvar2025divprune}, these methods commonly score visual tokens using text-to-vision attention magnitudes. 
Recent approaches further incorporate feature diversity, edge saliency, temporal feature changes, and moving averages of action-to-vision attention to guide token selection in sequential robotic observations~\cite{NEURIPS2025_3a2ef31a,li2025sp,wang2026specprunevla,liu2025bridging}.

Besides selection criteria, methods further differ in layer-aware treatment, as attention patterns shift across the language backbone's layers. Attention scores from deeper layers, which more directly reflect action-relevance, are used to guide token selection \cite{liu2025bridging,ma2026safe}.  However, since these scores are only available from the previous forward pass, relying on them degrades performance at low retention rates \cite{luo2026gate}. Other methods instead vary the pruning rate itself across layers, guided by heuristics such as attention entropy \cite{NEURIPS2025_f062da19,wang2026specprunevla,NEURIPS2025_3a2ef31a}.

Pruning is further regulated along the temporal dimension: full-context forward passes are executed at the beginning of the control sequence to fill up caches \cite{NEURIPS2025_f062da19,liu2025bridging} or the pruning rate is adjusted according to signals such as end-effector velocity \cite{ICLR2026_1223ca2b,li2025sp,wang2026specprunevla}. In both cases, these strategies rely on carefully tuned thresholds.

These layer and time-varying mechanisms come with an additional cost: accessing and modifying internal attention mechanisms are incompatible with fused-attention kernels such as FlashAttention \cite{flashattention} or graph-compiled inference, offsetting the very speedups they aim to provide. Taken together, training-free methods still struggle at high pruning rates, as they rely on a growing stack of heuristic-driven strategies that each require carefully tuned hyperparameters, rather than a signal tied directly to task performance.

\textit{Training-based} methods instead learn the pruning decision jointly with the VLA. To handle the non-differentiable hard pruning objective, LightVLA \cite{jiang2025better} trains the entire VLA stack using a Gumbel-softmax \cite{jang2017categorical} token selector. Grid-S \cite{feng2026see} replaces discrete token selection with a differentiable grid-sampling module fine-tuned end-to-end. Both report strong performance at aggressive pruning ratios, but the pruner and the backbone have to be re-trained, which can take multiple days even on 8 frontier GPUs \cite{kim2025fine} using parameter-efficient fine-tuning methods such as LoRA \cite{hu2022lora}. 

VLA-ACL sits in the gap between these two regimes: like the training-free methods, it keeps the VLA frozen, and it trains only a small patch scorer in just a few GPU hours; like the training-based methods, the pruning decisions are learned. A single hyperparameter $K$ controls how many visual tokens are kept per view, applied before tokens ever enter the language backbone. This mechanism remains constant at every control step, eliminating the need for temporal controllers and keeping VLA-ACL compatible with existing inference-acceleration techniques. Crucially, rather than scoring tokens through intermediate proxies, VLA-ACL learns this scoring function by optimizing action consistency between pruned and full-context forwards of the same frozen VLA, directly tying token selection to its effect on the predicted action.

\begin{figure*}[t]
\centering
\includegraphics[width=1.\textwidth]{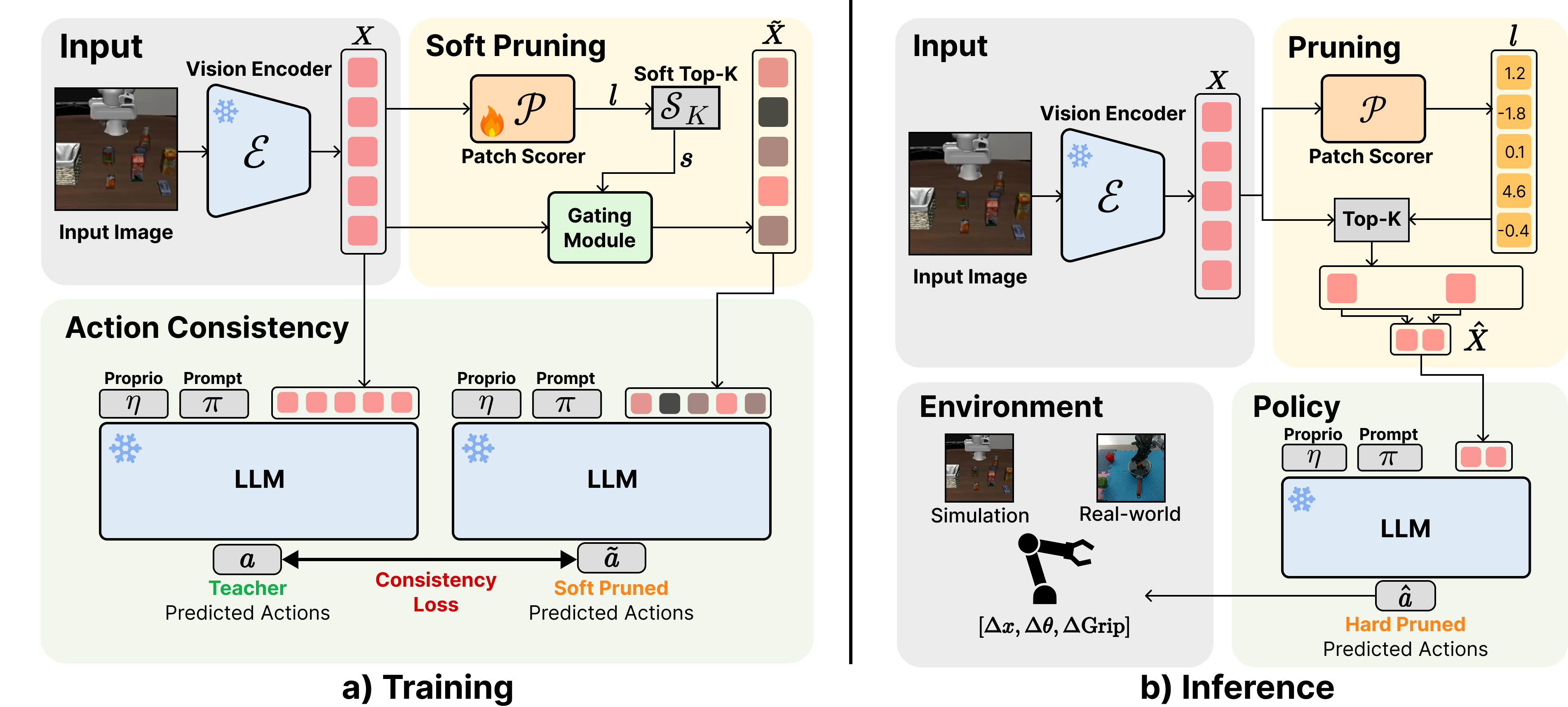}
\caption{Overview of VLA-ACL. Modules with a snowflake symbol are kept frozen, modules with a flame symbol receive gradient updates. \textbf{a)} During training, two forward passes of the frozen LLM are performed. On the left, the full-vision-context teacher action is generated. On the right, vision embeddings enter our differentiable soft pruning gate and produce the soft-pruned action. \textbf{b)} During inference, the hard pruning gate only lets through $K$ patches per view with the highest patch scores, obtained by the trained patch scorer. The pruned tokens never enter the VLA language backbone.}
\label{fig:full-pipeline}
\end{figure*}

\section{Method}
VLA-ACL learns a lightweight visual token pruning policy through action-level supervision while keeping the original VLA model frozen. A small Transformer-based~\cite{attention} patch scorer assigns an importance score to each visual token. At inference, the $K$ highest-scoring tokens are retained before entering the language backbone. 
During training, the scorer is optimized to maintain action consistency between pruned-context predictions and full-context teacher predictions, with ground-truth actions providing auxiliary supervision. A differentiable soft pruning module combines soft Top-$K$~\cite{lapsum2025} with a custom gating mechanism to enable gradient propagation to the scorer. Fig.~\ref{fig:full-pipeline} illustrates the training (a) and inference (b) pipelines.

Formally, let $\mathcal{E}$ denote the frozen vision encoder with embedding dimension $d$. Given a single-view image $I$, the encoder produces $N_e$ vision embeddings $\bm{X}=\mathcal{E}(I)\in\mathbb{R}^{N_e\times d}$. The trainable patch scorer $\mathcal{P}$ assigns each token a logit, yielding $\bm{l}=\mathcal{P}(\bm{X})\in\mathbb{R}^{N_e}$. For multi-view inputs (\emph{e.g.} wrist cameras), $\mathcal{P}$ jointly processes the concatenated embeddings from all views, while pruning and gating are applied independently to each view with a per-view token budget of $K$.

\subsection{Token-Pruning: Inference}\label{sec:inference}
At each control step, VLA-ACL retains the $K$ visual tokens with the highest patch scores $\bm{l}$.
Let $i_1,\dots,i_{N_e}$ index the tokens in descending score order:
\begin{align}
    \bm{l}_{i_1} \geq &\bm{l}_{i_2} \geq\dots\geq \bm{l}_{i_{N_e}},\\
    \text{Top-}K(\bm{X}; \bm{l}) &= [\bm{X}_{i_1},\bm{X}_{i_2},\dots,\bm{X}_{i_K}]\label{eq:topk-def}.
\end{align}
The selected embeddings pass through the frozen language backbone to predict actions conditioned on proprioceptive inputs $\eta$ and language instructions $\pi$:
\begin{align}
    \bm{l} &= \mathcal{P}(\bm{X})\in\mathbb{R}^{N_e}, \\
    \hat{\bm{X}} &= \text{Top-}K(\bm{X};\bm{l})\in\mathbb{R}^{K\times d},\\
    \hat{\bm{a}} &= \text{LLM}(\hat{\bm{X}},\eta,\pi).
    \label{eq:hard-pruning-action}
\end{align}
Discarded tokens are excluded from this computation. Each retained token preserves its original RoPE position index ~\cite{su2024roformer}, maintaining the positional relationships of the full visual sequence.

\subsection{Action Consistency Learning}\label{sec:training}
Action Consistency Learning trains the patch scorer to preserve the frozen VLA's action predictions under visual token pruning. The full-context VLA provides teacher supervision, complemented by ground-truth actions. However, hard Top-$K$ selection is piecewise constant with respect to the patch logits and has zero gradients almost everywhere. We therefore introduce a differentiable training surrogate that assigns soft selection scores and gates visual embeddings, allowing action-level supervision to reach the scorer through the frozen VLA.

\textbf{Soft token selection.}
We use soft Top-$K$, a differentiable relaxation~\cite{lapsum2025} that maps patch logits $\bm{l}$ to continuous saliency scores.
Formally, the operator $\mathcal{S}_K$ determines an adaptive threshold $\theta_K$ such that the total selection mass equals $K$, assigning each token a score based on its logit's scaled distance from this threshold:
\begin{gather}
    \bm{s} = \mathcal{S}_K(\bm{l})\in[0,1]^{N_e},\\
    \bm{s}_i = \mathrm{Lap}\left(\frac{\bm{l}_i-\theta_K}{\alpha}\right),
    \qquad\sum_{i=1}^{N_e}s_i = K.\label{eq:soft-topk}
\end{gather}
where $\mathrm{Lap}$ denotes the standard Laplace cumulative distribution function and $\alpha>0$ controls selection softness. As $\alpha\to0^+$, $\mathcal{S}_K$ approaches exact binary Top-$K$. We anneal $\alpha$ with a cosine schedule to progressively sharpen selection while keeping the budget $K$ fixed.

\begin{figure}[tbp]
\vspace{4pt}
\centering
\includegraphics[width=0.48\textwidth]{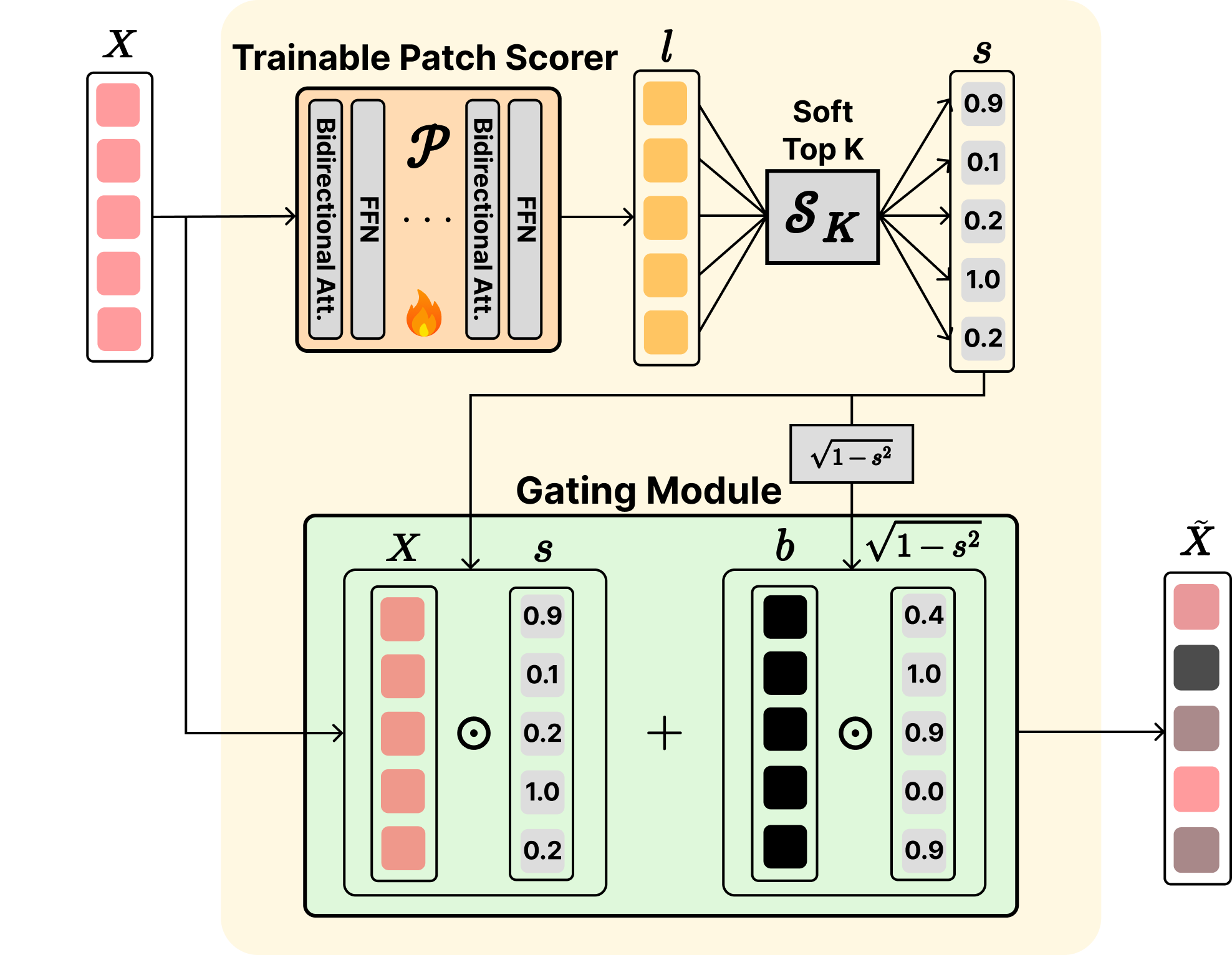}
\caption{The soft pruning gate consists of the trainable patch scorer, which outputs a logit for each vision patch. The logits pass through a soft Top-$K$ function, which maps logits to $[0,1]$ scores summing to $K$. These scores regulate the information content per patch inside the gating module, where patches with low scores are perturbed using embeddings $\bm{b}$ of a pitch-black image.}
\label{fig:soft-gate}
\end{figure}

\textbf{Differentiable gating.}
We use the selection scores $\bm{s}$ from Eq.~\ref{eq:soft-topk} to construct soft-pruned visual embeddings $\Tilde{\bm{X}}$ without discretely removing tokens. Let $I_{\mathrm{black}}$ denote a black image processed with the same input normalization as the original observations. The frozen vision encoder produces black image embeddings $\bm{b}$, which are combined with the original embeddings through a differentiable gate:
\begin{align}
    \bm{b} &= \mathcal{E}(I_{\mathrm{black}})\in\mathbb{R}^{N_e\times d},\label{eq:black-patch} \\
    \Tilde{\bm{X}}_i &=\bm{X}_i*\bm{s}_i + \bm{b}_i *\sqrt{1-\bm{s}_i^2}\label{eq:interpolation}.
\end{align}
The saliency score $\bm{s}_i$ determines how much of the original vision embedding $\bm{X}_i$'s information survives: for $\bm{s}_i$ close to 1, $\bm{X}_i$ is fully preserved, while $\bm{s}_i\to0$ replaces the patch entirely with the black patch embedding. 
Using the encoder's representation of a black image keeps the replacement in-distribution for the frozen backbone, unlike zero vectors or random noise.
The resulting soft-pruned embeddings $\Tilde{\bm{X}}$ serve as a differentiable surrogate for the hard-pruned $\hat{\bm{X}}$, allowing action-level gradients to reach the patch scorer $\mathcal{P}$. The soft pruning gate is shown in Fig.~\ref{fig:soft-gate}.

\textbf{Action-level supervision.}
Both original $\bm{X}$ and soft-pruned vision embeddings $\Tilde{\bm{X}}$ pass separately through the frozen LLM, together with proprioception data $\eta$ and language instructions $\pi$:
\begin{align}
        \Tilde{\bm{a}} &= \text{LLM}(\Tilde{\bm{X}}, \eta, \pi),\label{eq:soft-prune-action}\\
    \bm{a} &= \text{LLM}(\bm{X}, \eta,\pi).
\end{align}
With the obtained teacher action $\bm{a}$ and soft-pruned action $\Tilde{\bm{a}}$, we optimize the action consistency loss $\mathcal L_{\text{acl}}$ with auxiliary ground-truth supervision $\bm{a}_{\text{gt}}$:
\begin{equation}
    \mathcal L_{\text{acl}} = \lambda ||\Tilde{\bm{a}}-\bm{a}||_1+\mu ||\Tilde{\bm{a}}-\bm{a}_{\text{gt}}||_1, \label{eq:hybrid_loss}
\end{equation}
where $\lambda$ and $\mu$ weight the consistency and ground-truth terms, respectively.
The consistency term encourages token selection to preserve the full-context teacher's actions, while the auxiliary term anchors predictions to demonstrated actions when the teacher is inaccurate.
Section~\ref{sec:ablation} provides more detailed analysis and ablations. Gradients of $\mathcal L_{\text{acl}}$ propagate through the frozen VLA and differentiable gate to update only the patch scorer.

We optimize this objective with a fixed budget $K$, analogous to Top-$K$ sparse autoencoders~\cite{ICLR2025_42ef3308}, while annealing progressively sharpens selection. Unlike an $\ell_1$ sparsity penalty, the fixed budget directly controls the inference pruning ratio without an additional regularization hyperparameter.

\begin{table*}[tbp]
\vspace*{4pt}
\caption{Results of \textbf{VLA-ACL} on the LIBERO suite compared with OpenVLA-OFT and other frozen-VLA pruning and caching methods.}
\label{tab:main_libero_results}
\begin{center}
\resizebox{0.85\textwidth}{!}{
\begin{tabular}{lccccccc}
\toprule
\multirow{2}{*}{Method} & \multicolumn{5}{c}{Success Rate $\uparrow$} & \multirow{2}{*}{TFLOPs} & \multirow{2}{*}{Latency (ms)} \\
\cmidrule(lr){2-6}
& Spatial & Object & Goal & Long & Average & & \\
\midrule

OpenVLA-OFT & 97.8 & 97.6 & 97.6 & 94.2 & 96.8 & 4.013 & 63.49 \\
+ VLA-Cache & 98.3 & 97.5 & 98.3 & 95.4 & 97.4 & 3.097 & 66.79 \\
+ VLA-ADP & 99.4 & 98.0 & 96.4 & 91.2 & 96.3 & 3.139 & 58.26 \\
+ SAFE-Pruner & 98.0 & 98.0 & 96.2 & 93.4 & 96.4 & 1.946 & 56.61 \\
+ SpecPrune-VLA & 97.4 & 95.8 &  97.7 &  93.4 & 96.1 & 1.726 & 60.36\\
+ VLA-Pruner & 93.5 & 96.2 & 95.2 & 90.2 & 93.8 & 1.420 & 63.11 \\

\midrule
+ \textbf{VLA-ACL} (K=32) & 98.2 & 98.2 & 96.6 & 88.7 & 95.4 & 0.991 & 42.30 \\
+ \textbf{VLA-ACL} (K=64) & 98.7 & 98.4 & 96.5 & 94.2 & 97.0 & 1.412 & 46.82 \\

\bottomrule
\end{tabular}
}
\end{center}
\end{table*}
\begin{figure*}[tbp]
\centering
\includegraphics[width=0.95\textwidth]{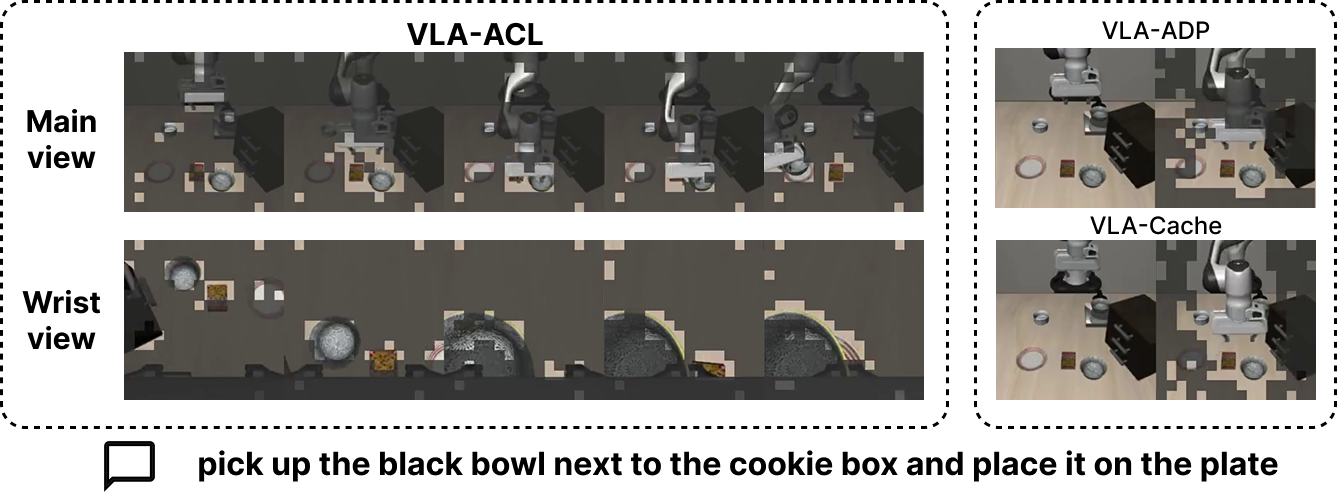}
\caption{Visualization of pruning/caching policies on LIBERO-Spatial. \textcolor{Gray}{Grayed} patches are pruned (VLA-ACL, VLA-ADP) or used as previously cached representations (VLA-Cache). VLA-ADP and VLA-Cache both rely on full-context forwards during inference, while \textbf{VLA-ACL} sustains a much more aggressive pruning schedule at every control step, enabling lower latency while retaining similar performance across all LIBERO suites.}
\label{fig:libero-pruning}
\end{figure*}

\section{Experiments}\label{sec:experiments}

We evaluate VLA-ACL on the LIBERO \cite{NEURIPS2023_8c3c6668} simulation benchmark and on real-world manipulation tasks. Furthermore, we conduct ablation studies examining our training objective, token selection strategy, and gating mechanism. We compare against other frozen-VLA pruning and caching methods built on OpenVLA-OFT. To test generality beyond this backbone, we additionally apply VLA-ACL to $\pi_{0.5}$ (Appendix~\ref{sec:appendix-pi}). All experiments and measurements are conducted using a single NVIDIA A100 GPU.

\subsection{Training}

The patch scoring network is a 5-layer bidirectional Transformer that operates on the vision embeddings produced by the vision encoder for each camera view. For the LIBERO suite, we use two camera views, each represented by 256 visual tokens. For two real-world tasks, we use three camera views: two wrist cameras and one static camera.
For each task, we train and evaluate the model at $K=32$ and $K=64$, corresponding to a token pruning rate of 87.5\% and 75\%, respectively. These values span an aggressive and a moderate pruning regime, trading off performance against efficiency.

We use the following training hyperparameters uniformly across all LIBERO and real-world tasks, and for both VLA backbones (Appendix~\ref{sec:appendix-impl}). We set $\lambda=0.5$ and $\mu=0.5$, equally weighting action consistency to the teacher and to the ground-truth demonstration. $\alpha$ is annealed from $2.0$ to $0.1$ across 5000 training steps, closing the gap between soft and hard pruning regimes over the course of training. We optimize with \texttt{AdamW} \cite{loshchilov2018decoupled}, using a learning rate of $1\times10^{-4}$ and a batch size of 8.

\subsection{Simulation Experiments}
\textbf{Experiment setup.} We evaluate on LIBERO's four suites (Spatial, Object, Goal, Long), testing spatial reasoning, object recognition, goal-directed behavior and long-horizon planning. We adopt the LIBERO evaluation settings of OpenVLA-OFT, consistent with other baselines, where each control step predicts a chunk of eight consecutive actions. Success rates of VLA-ACL are averaged across three random seeds with 50 trials per task, yielding 1500 episodes per task suite. Baseline success rates are taken directly from the corresponding publications. 
We restrict comparisons to methods that keep the underlying VLA frozen, matching VLA-ACL’s backbone setting and enabling a controlled comparison of pruning strategies.
Latency and FLOPs of all methods are computed following the protocol of VLA-Cache \cite{NEURIPS2025_f062da19}, measuring the TFLOPs and inference time of the LLM. These efficiency metrics are widely adopted across VLA acceleration methods.

\textbf{Experiment results.} Table~\ref{tab:main_libero_results} summarizes results on the LIBERO benchmark. At $K\!=\!32$, VLA-ACL reduces FLOPs by 75\% and achieves a 1.5$\times$ speed-up relative to the OpenVLA-OFT baseline, at the cost of only a marginal decrease ($-1.4\%$) in average success rate. At $K\!=\!64$, our method reduces FLOPs by 65\% and achieves a 1.4$\times$ speed-up, while slightly exceeding the baseline's average ($+0.2\%$). The two configurations differ most substantially on LIBERO-Long, where long-horizon trajectories are more sensitive to the reduced visual context available at $K\!=\!32$. Relative to other methods, VLA-ACL maintains strong performance under aggressive pruning and achieves the lowest latency. VLA-Cache, SpecPrune-VLA \cite{wang2026specprunevla}, and VLA-Pruner \cite{liu2025bridging} rely on caching internal activations of the LLM, breaking compatibility with FlashAttention \cite{flashattention} and requiring extra GPU time to allocate a \texttt{DynamicCache}. Consequently, latency gains from FLOP savings are offset: despite a reduction in FLOPs, VLA-Cache runs slower than base OpenVLA-OFT across the GPU configurations we tested.

VLA-ACL also transfers across VLA backbones: applied to $\pi_{0.5}$~\cite{Intelligence202505AV} with identical hyperparameters and a single scorer trained jointly on all four suites, it ($K\!=\!64$) reduces TFLOPs by 67\% and pre-fill latency by $2.2\times$ at 96.1\% average success ($-0.8$ vs.\ the unpruned model). It outperforms SAFE-Pruner \cite{ma2026safe} in success rate ($+0.6$) with less than half of its TFLOPs (Appendix~\ref{sec:appendix-pi}) while retaining \texttt{torch.compile}.

\textbf{Visualizations.} Fig.~\ref{fig:libero-pruning} visualizes VLA-ACL's pruning policy across an entire LIBERO-Spatial episode, compared to VLA-ADP \cite{ICLR2026_1223ca2b} and VLA-Cache. VLA-ACL prunes every control step (including the first frame) at a constant rate without temporal controllers or falling back to full context, as opposed to the other methods. Our pruning policy reveals which patches contribute most to the action produced by the frozen VLA, whereas training-free methods prune based on heuristics such as internal attention values that only indirectly reflect the token's attribution toward the VLA action.

\subsection{Real-world experiments}

\begin{figure*}[tbp]
\centering
\includegraphics[width=\textwidth]{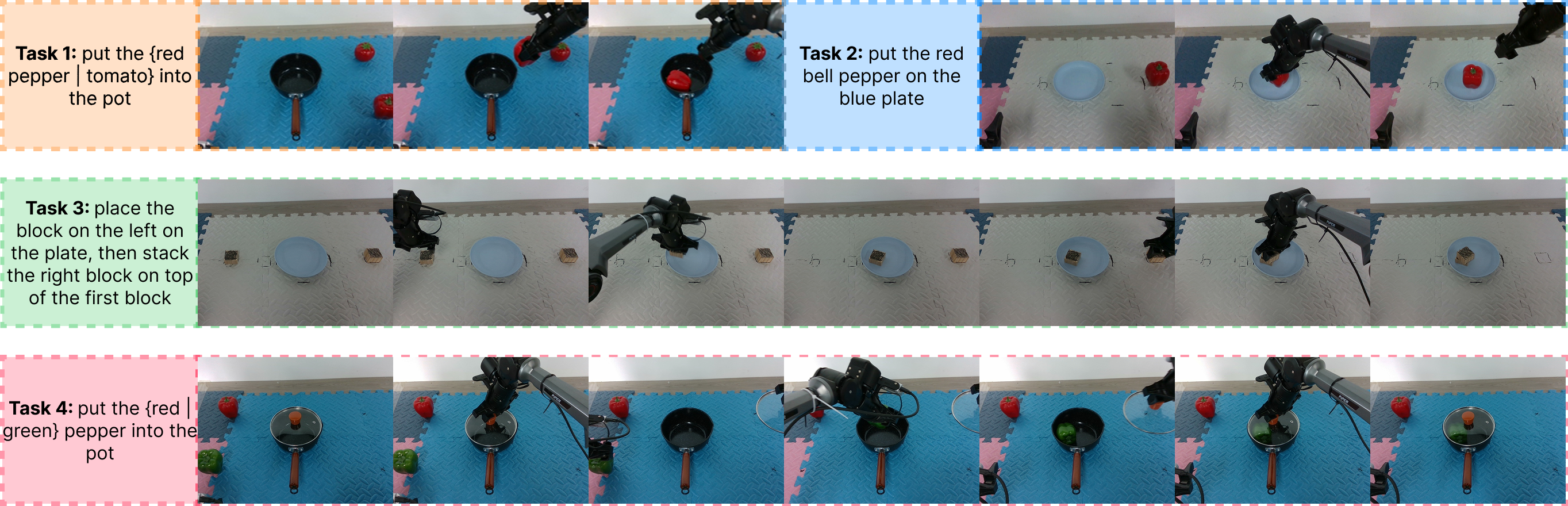}
\caption{Real-world experiment task trajectories captured with RealSense D435 cameras.}
\label{fig:rl-exp}
\end{figure*}

\textbf{Experiment setup.} We evaluate on an ALOHA-style~\cite{Zhao2023LearningFB} bimanual teleoperation platform (AgileX PiPER-X). Following OpenVLA-OFT's real-world experiment protocol, we set the action chunk size to 25, corresponding to one second of execution in real time. Each of the four tasks is trained on 50--100 expert demonstrations. Tasks 1 and 4 evaluate instruction following and precise perception; task 4 is additionally a long-horizon task comprising multiple subtasks (lifting the pot lid, picking the correct vegetable, placing the lid back). For both tasks, we vary object positions and the corresponding language instruction across 4 configurations. Tasks 2 and 3 assess spatial understanding and goal-directed behavior, with task 3 featuring multiple subtasks. Tasks 2 and 3 use three camera views and both arms, while the remaining tasks use two and only the right arm. Fig.~\ref{fig:rl-exp} shows the main camera view along each task trajectory and task instructions.

\textbf{Experiment results.} Table~\ref{tab:real-world-results} reports results on the real-world tasks, where VLA-ACL is evaluated at $K=64$. We run 10 trials per configuration for tasks 1 and 4 (40 trials per task in total) and 20 trials for tasks 2 and 3, respectively. VLA-ACL trails the OpenVLA-OFT baseline by a single failed trial on tasks 1 and 3, while matching or exceeding it on the other two tasks. Notably, both models excel at simple pick-place tasks (tasks 2 and 3) but face slight difficulties with fine-grained visual perception in task 1, and significant challenges when combining instruction following with long-horizon execution in task 4. In total, we measure a FLOPs reduction of more than 50\% and a speed-up of around 1.5$\times$. Averaged across all four tasks, VLA-ACL delivers performance comparable to full-context baselines in real-world settings while enabling substantially faster inference.

\begin{table*}[tbp]
\vspace*{4pt}
\caption{Results of \textbf{VLA-ACL} at $K=64$ on real-world tasks compared with OpenVLA-OFT.}
\label{tab:real-world-results}
\begin{center}
\resizebox{0.75\textwidth}{!}{
\begin{tabular}{lccccccc}
\toprule
\multirow{2}{*}{Method} & \multicolumn{5}{c}{Success Rate $\uparrow$} & \multirow{2}{*}{TFLOPs} & \multirow{2}{*}{Latency (ms)} \\
\cmidrule(lr){2-6}
& Task 1 & Task 2 & Task 3 & Task 4 & Average & & \\
\midrule
OpenVLA-OFT & 75 & 95 & 85 & 40 & 73.75 & 6.257 & 91.5 \\
+ \textbf{VLA-ACL} & 72.5 & 95 & 80 & 42.5 & 72.5 & 2.953 & 62.1 \\
\bottomrule
\end{tabular}}
\end{center}
\end{table*}

\subsection{Ablation study}\label{sec:ablation}
The following ablation studies are conducted with OpenVLA-OFT.

\begin{figure}[htbp]
\centering
\includegraphics[width=0.45\textwidth]{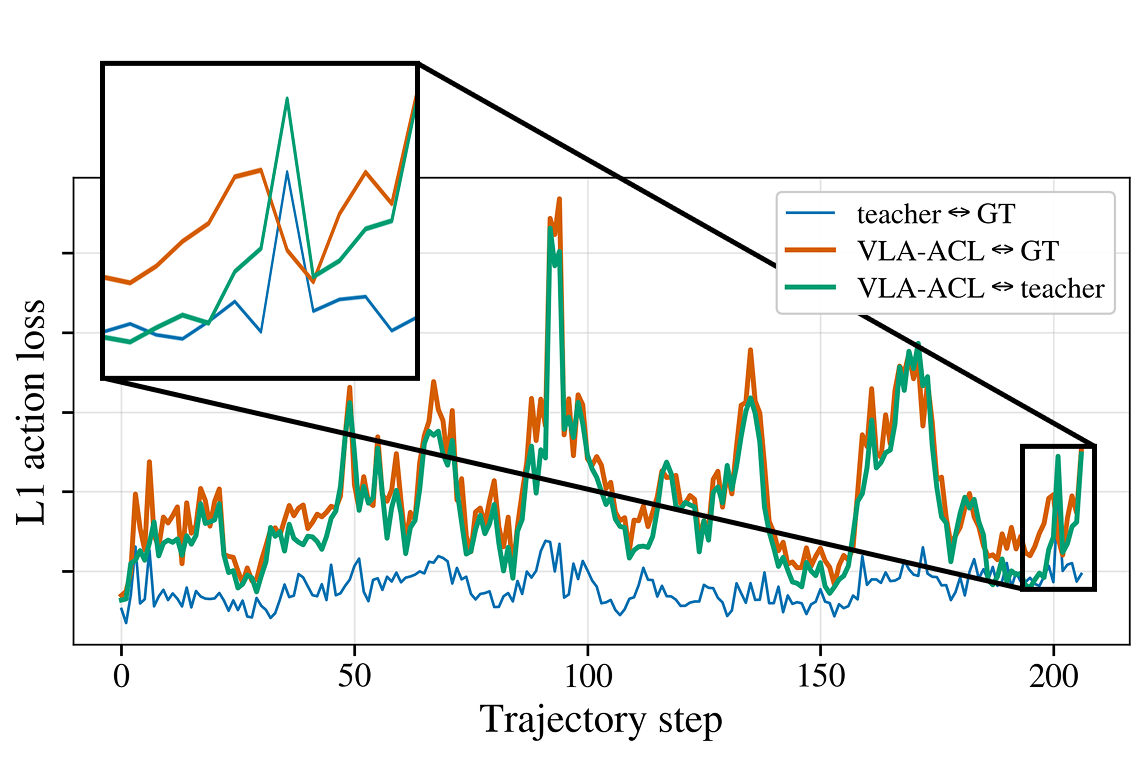}
\caption{$\ell_1$ distance between VLA-ACL ($K=64$), teacher, and ground-truth (GT) actions over a LIBERO-Long training episode.}
\label{fig:action-loss-episode}
\end{figure}

\textbf{Action consistency objective.} Table~\ref{tab:objective_ablation} shows that our action consistency objective with auxiliary ground-truth supervision ($\lambda=0.5,\mu=0.5$) outperforms both the teacher-only and ground-truth (GT) only variants across $K$ configurations. Because the fine-tuned OpenVLA-OFT models are trained to an action loss below $0.01$, teacher and ground-truth actions coincide closely in most frames. However, this gap can nonetheless spike within an episode; thus, including ground-truth actions in the training objective supplies additional supervision signal. To examine this effect directly, Fig.~\ref{fig:action-loss-episode} tracks the $\ell_1$ distance of the VLA-ACL predicted action to the full-context teacher and to the GT action over a single training episode. At certain steps, VLA-ACL$\leftrightarrow$GT not only falls below VLA-ACL$\leftrightarrow$teacher -- demonstrating the effect of the auxiliary supervision -- but also falls below the teacher$\leftrightarrow$GT itself. Despite VLA-ACL only operating on the frozen VLA, this indicates that the learned pruning policy can push the predicted pruned action closer to the ground-truth signal than the unpruned teacher.

\begin{table}[htbp]
\caption{LIBERO success rates of different training objectives.}
\label{tab:objective_ablation}
\begin{center}
\begin{tabular}{cc|c|c}
\toprule
Teacher & GT & K & Success Rate $\uparrow$ \\
\midrule
\cmark& \xmark& \multirow{3}{*}{32}&  95.2 \\
\xmark& \cmark& & 95.1 \\
\cmark& \cmark& &\textbf{95.4} \\
\midrule
\cmark& \xmark&  \multirow{3}{*}{64}& 96.4 \\
\xmark& \cmark&&  95.9 \\
\cmark& \cmark& & \textbf{97.0} \\
\bottomrule
\end{tabular}
\end{center}
\end{table}

\begin{table*}[t]
\vspace*{4pt}
\caption{Comparison of selection methods at equal budget. Action loss denotes $\mathcal{L}_{acl}$ with $\lambda=0.5$ and $\mu=0.5$. K denotes per-view visual token budget.}
\label{tab:equal_budget}
\begin{center}
\resizebox{0.9\textwidth}{!}{
\begin{tabular}{lccccccc}
\toprule
\multirow{2}{*}{Method} & \multirow{2}{*}{K} & \multirow{2}{*}{Pruning layer} & \multicolumn{5}{c}{Success Rate $\uparrow$ / \textit{Action Loss} $\downarrow$} \\
\cmidrule(lr){4-8}
& & & Spatial & Object & Goal & Long & Average \\
\midrule

Attention & \multirow{4}{*}{32} & pre & 61.8 / \textit{0.127} & 55.0 / \textit{0.131} & 59.0 / \textit{0.096} & 3.2 / \textit{0.122} & 44.8 / \textit{0.119} \\
Attention + RoPE & & pre & 96.6 / \textit{0.061} & 95.0 / \textit{0.065} & 96.6 / \textit{0.058} & 69.6 / \textit{0.064} & 89.5 / \textit{0.062} \\
VLA-Pruner & & 3 & 88.1 / \textit{0.064} & 87.6 / \textit{0.100} & 84.9 / \textit{0.070} & 68.8 / \textit{0.052} & 82.4 / \textit{0.071} \\
VLA-ACL & & pre & 98.2 / \textit{0.033} & 98.2 / \textit{0.043} & 96.6 / \textit{0.047} & 88.7 / \textit{0.037} & 95.4 / \textit{0.040} \\
\midrule
Attention & \multirow{4}{*}{64} & pre & 77.4 / \textit{0.110} & 59.8 / \textit{0.111} & 76.6 / \textit{0.079} & 33.2 / \textit{0.096} & 61.8 / \textit{0.099} \\
Attention + RoPE & & pre & 97.4 / \textit{0.028} & 98.4 / \textit{0.031} & 96.0 / \textit{0.036} & 90.2 / \textit{0.033} & 95.5 / \textit{0.032} \\
VLA-Pruner & & 3 & 93.5 / \textit{0.054} & 96.2 / \textit{0.079} & 95.2 / \textit{0.063} & 90.2 / \textit{0.040} & 93.8 / \textit{0.059} \\
VLA-ACL & & pre & 98.7 / \textit{0.016} & 98.4 / \textit{0.022} & 96.5 / \textit{0.034} & 94.2 / \textit{0.019} & 97.0 / \textit{0.023} \\

\bottomrule
\end{tabular}
}
\end{center}
\end{table*}

\textbf{Selection method.} To validate that our action-consistency-based selection outperforms the heuristics used by frozen-VLA pruning methods, we compare against three baselines on LIBERO (Table~\ref{tab:equal_budget}). \textit{Attention} denotes the text-to-vision attention heuristic introduced by FastV \cite{chen2024image} and adopted by VLA-ADP \cite{ICLR2026_1223ca2b}: first-layer attention scores are used to prune non-salient vision tokens before the forward pass through the language backbone at every control step. \textit{Attention + RoPE} augments this baseline with RoPE~\cite{su2024roformer} positions taken from each token's original index rather than its position in the pruned sequence, preserving spatial context for the frozen VLA. \textit{VLA-Pruner} combines layer-3 attention scores with an estimate of action-to-vision attention, pruning vision tokens inside the LLM after the third layer. We report success rate and action consistency loss (Eq.~\ref{eq:hybrid_loss}) for each method across per-view token budgets $K$.

The results show that plain text-to-vision attention performs poorly, collapsing in low token retention settings on LIBERO-Long ($\sim3\%$ SR). Adding RoPE positions from the original sequence recovers strong success rates and lower action loss, even surpassing VLA-Pruner. VLA-ACL nonetheless outperforms all baselines in both action loss and success rate, demonstrating that token selection under action-consistency learning retains the most performance at equal token budgets. Success rate and action loss are broadly correlated across methods, although the relationship saturates on some suites: on LIBERO-Goal at $K=32$, VLA-ACL attains a lower action loss than \textit{Attention + RoPE} ($0.047$ vs. $0.058$) at equal success rates ($96.6\%$ for both), suggesting performance is near ceiling on that suite.

\begin{figure}[htbp]
\centering
\includegraphics[width=0.45\textwidth]{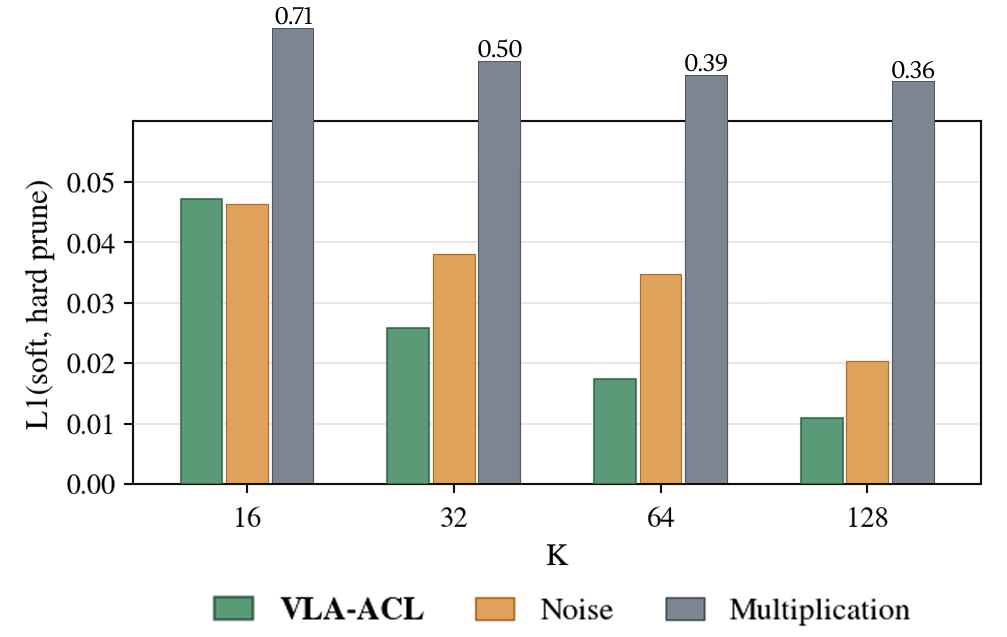}
\caption{$\ell_1$ between soft-pruned actions of different gating modes vs. hard-pruned actions at different $K$. Smaller is better.}
\label{fig:soft-vs-hard}
\end{figure}

\textbf{Gating mode.} We show that gating with black-patch embeddings (Eq.~\ref{eq:black-patch}) effectively restricts information during soft pruning, mimicking the inference-time hard pruning regime. Since the soft-pruned action $\Tilde{\bm{a}}$ should closely track the hard-pruned action $\hat{\bm{a}}$ — obtained via hard Top-$K$ token selection — to prevent a training-deployment gap, we quantify this via the soft-to-hard $\ell_1$ action gap. A smaller difference indicates that soft pruning better emulates hard pruning. We compare VLA-ACL's gating against noise gating, as used by DiffPrune~\cite{he2026beyond}, which substitutes Gaussian noise for black-patch embeddings. Their approach requires training an additional denoiser, since heavily noised patches fall outside the training distribution. Moreover, we compare with multiplicative gating, which simply scales the patches $\bm{X}$ directly by their soft Top-$K$ output $\bm{s}$ and drops the $\sqrt{1-s^2}$ term in Eq.~\ref{eq:interpolation}. As shown in Fig.~\ref{fig:soft-vs-hard} across $K$ values at $\alpha=0.1$, multiplication maintains a persistently high action loss, while VLA-ACL's black-patch gating yields a significantly smaller soft-to-hard gap than noise gating for $K\geq32$. This confirms that our soft pruning scheme closely mirrors the hard Top-$K$-pruned action used at inference, making it an accurate optimization proxy during training.

\section{Conclusion}
We introduced VLA-ACL, a visual token pruning framework for frozen VLAs that grounds token selection in action consistency: a lightweight patch scorer is trained so that actions produced from a pruned visual context match those of the same model with full context, with ground-truth demonstrations as auxiliary supervision. On LIBERO across two VLA backbones and on real-world tasks, VLA-ACL prunes up to 87.5\% of visual tokens and cuts FLOPs by up to 75\% while maintaining competitive success rates, at a training cost of only a few GPU hours, extending the Pareto frontier of FLOPs and latency among frozen-VLA visual token pruning methods.




\bibliographystyle{IEEEtran}
\bibliography{references}

\clearpage
\appendix

\begin{table*}
\caption{Results of \textbf{VLA-ACL} on the LIBERO suite compared with $\pi_{0.5}$ and other frozen-VLA pruning and caching methods. Success rates and TFLOPs of the baselines are taken from the results reported by SAFE-Pruner~\cite{ma2026safe}; their latency cannot be measured as the code is not public.}
\label{tab:libero_results_pi}
\begin{center}
\resizebox{0.9\textwidth}{!}{
\begin{tabular}{lccccccc}
\toprule
\multirow{2}{*}{Method} & \multicolumn{5}{c}{Success Rate $\uparrow$} & \multirow{2}{*}{TFLOPs} & \multirow{2}{*}{Latency (ms)} \\
\cmidrule(lr){2-6}
& Spatial & Object & Goal & Long & Average & & \\
\midrule

$\pi_{0.5}$ & 98.8 & 98.2 & 98.0 & 92.4 & 96.9 & 2.115 & 25.9 \\
+ VLA-Cache & 94.8 & 96.0 & 96.4 & 88.2 & 93.9 & 1.632 & - \\
+ VLA-Pruner & 95.8 & 97.2 & 95.6 & 88.0 & 94.2 & 1.583 & - \\
+ SAFE-Pruner & 97.0 & 98.8 & 96.0 & 90.2 & 95.5 & 1.482 & - \\

\midrule
+ \textbf{VLA-ACL} (K=64) & 97.4 & 98.0 & 96.6 & 92.3 & 96.1 & 0.701 & 11.6 \\

\bottomrule
\end{tabular}
}
\end{center}
\end{table*}

\subsection{Implementation Details}\label{sec:appendix-impl}
\textbf{OpenVLA-OFT.} We use the OpenVLA-OFT model fine-tuned per LIBERO suite, like the methods we compare with, and train a patch scorer for each model on the trajectories of the respective suite. For the real-world experiments, we likewise fine-tune one base policy per task, following the OpenVLA-OFT specification. All ablation studies in Section~\ref{sec:ablation} are conducted with OpenVLA-OFT.

{$\bm{\pi_{0.5}}$.} We further train VLA-ACL on $\pi_{0.5}$, following the same action consistency objective as in Eq.~\ref{eq:hybrid_loss}. Since $\pi_{0.5}$ produces actions through flow matching~\cite{lipman2023flow} rather than direct regression, we adopt the base model's training procedure for the patch scorer, sampling a noise level per batch and perturbing the ground-truth actions accordingly. The consistency term is then computed between the velocity predicted from the pruned visual context and that of the full-context teacher at the same noise level, while the auxiliary term supervises the predicted velocity with the flow-matching target. In contrast to OpenVLA-OFT, $\pi_{0.5}$ has a single model for all LIBERO suites. We therefore train one patch scorer on top of it jointly on all four suites and evaluate it on each, so the scorer generalizes across suites without per-suite specialization. All training hyperparameters ($\lambda$, $\mu$, the $\alpha$ schedule, learning rate, batch size and training steps) are identical to those used for OpenVLA-OFT (Section~\ref{sec:experiments}), without any backbone-specific tuning, showing that VLA-ACL generalizes across VLA backbones.

\textbf{Efficiency measurement.} For $\pi_{0.5}$, latency and TFLOPs follow the implementation of SAFE-Pruner~\cite{ma2026safe}, where only the LLM pre-fill is measured. The success rates and TFLOPs of VLA-Cache, VLA-Pruner, and SAFE-Pruner are taken from the SAFE-Pruner publication. We cannot measure their latency, as their code for this model is not public. Their temporal and per-layer pruning control, however, prevents the use of \texttt{torch.compile}, which we expect to slow down inference. Training and evaluation of all models are performed on a single NVIDIA A100 GPU.

\subsection{More results}\label{sec:appendix-pi}
Results on the $\pi_{0.5}$ model are summarized in Table~\ref{tab:libero_results_pi}. VLA-ACL at $K=64$ achieves the best average success rate among the frozen-VLA pruning baselines at the lowest computational budget, with a $67\%$ reduction in TFLOPs and a $2.2\times$ speedup in pre-fill latency relative to the base model. Compared to the strongest baseline, SAFE-Pruner, VLA-ACL uses less than half of the TFLOPs ($0.701$ vs.\ $1.482$) while exceeding its average success rate ($96.1\%$ vs.\ $95.5\%$), trailing the unpruned $\pi_{0.5}$ by only $0.8$ points. The gap is concentrated on LIBERO-Object, where SAFE-Pruner is ahead ($98.8\%$ vs.\ $98.0\%$). Since the baseline implementations are not available to us, we cannot measure their latency directly, but we expect the latency advantage measured on OpenVLA-OFT in Table~\ref{tab:main_libero_results} to carry over to this backbone. These results further show that a single patch scorer trained jointly across all four suites performs well on each of them.

\subsection{Real-world experiment setup}\label{sec:appendix-rw}
\begin{figure}[htbp]
\centering
\includegraphics[width=0.45\textwidth]{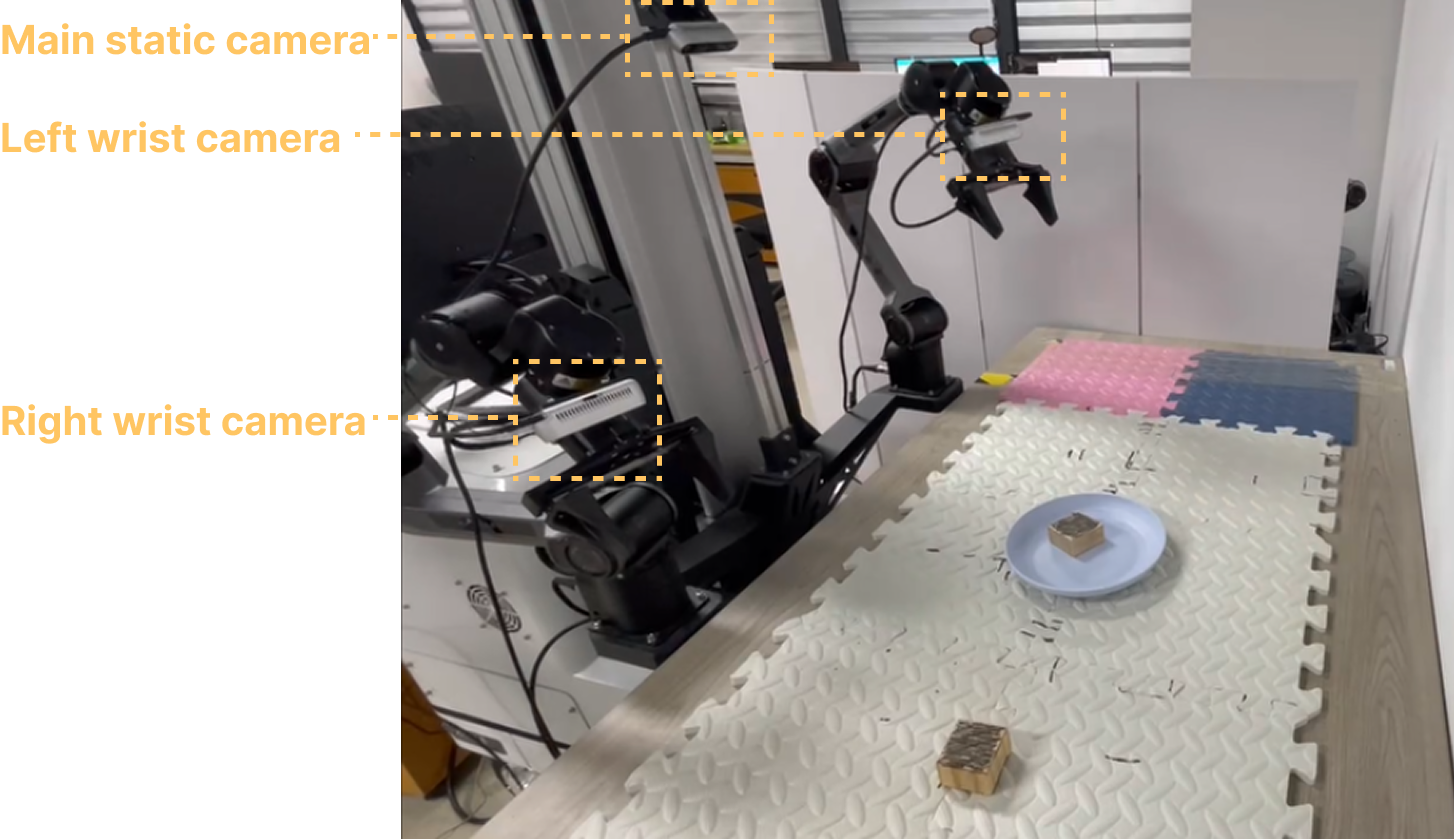}
\caption{Our real-world platform: a bimanual AgileX PiPER-X with a main static camera and left and right wrist cameras (RealSense D435).}
\label{fig:rl-setup}
\end{figure}

Fig.~\ref{fig:rl-setup} shows our bimanual robot platform (AgileX PiPER-X) with one static and two wrist-mounted RealSense D435 cameras. We collect 50--100 expert demonstrations for each of the four real-world tasks using the ALOHA teleoperation platform. Subsequently, we use the trajectory data to fine-tune OpenVLA-OFT using imitation learning.

\addtolength{\textheight}{-12cm}   




\end{document}